\documentclass{article}
\usepackage{graphicx}
\usepackage{amssymb,amsmath}
\usepackage{multicol}
\usepackage{eqnarray}
\usepackage{caption}
\usepackage{subcaption}
\usepackage{textcomp}
\usepackage{multirow}
\usepackage{float}
\usepackage{bm}
\usepackage{color}
\usepackage{epstopdf}
\usepackage{hyperref}
\usepackage{geometry}
\usepackage{cite}
\begin{document}
\title{A Generalized Strong--Inversion CMOS Circuitry for Neuromorphic Applications} 
\date{} 
\author{Hamid Soleimani and Emmanuel.~M.~Drakakis}
\maketitle{} 
\begin{abstract}
It has always been a challenge in neuromorphic field to systematically translate biological models into analog electronic circuitry. In this paper, a generalized circuit design platform is introduced where biological models can be conveniently implemented using CMOS circuitry operating in strong--inversion. The application of the method is demonstrated by synthesizing a relatively complex two--dimensional (2--D) nonlinear neuron model. The validity of our approach is verified by nominal simulated results with realistic process parameters from the commercially available AMS 0.35 $\mu m$ technology. The circuit simulation results exhibit regular spiking response in good agreement with their mathematical counterpart.
\end{abstract}

\section{Introduction}
Researchers in the neuromorphic community intend to mimic the neuro-biological structures in the nervous system using electronic circuitry. To do so different approaches have been developed so far:

\begin{enumerate}
\item Special purpose computing architectures have been developed to simulate complex biological networks via special software tools \cite{1,2,3,4,5}. Even though these systems are biologically plausible and flexible with remarkably high performance thanks to their massively parallel architecture, they run on bulky and power-hungry workstations with relatively high cost and long development time.

\item Digital platforms are good candidates nowadays for implementing such biological and bio-inspired systems. Most digital approaches \cite{6,7,8,9,10,11,12,13,14,15}, use digital computational units to implement the mathematical equations codifying the behavior of biological intra/extracellular dynamics. Such an approach can be either implemented on FPGAs or custom ICs, with FPGAs providing lower development time and more configurability. Generally, a digital platform benefits from high reconfigurability, short development time, notable reliability and immunity to device mismatch. Although, the digital platform's silicon area and power consumption is comparatively high compared to its analog counterpart.

\item Analog CMOS platforms are considered to be the main choice for direct implementation of intra-- and extracellular biological dynamics \cite{16,17,18,19,20,21,22,23,24}. This approach is very power efficient, however, model development and adjustment is generally challenging. Moreover, since the non--linear functions in the target models are directly synthesized by exploiting the inherent non--linearity of the circuit components, very good layout is imperative in order for the resulting topology not to suffer from the variability and mismatch particularly CMOS circuits operating in subthreshold. 
\end{enumerate}

\par To address the challenges explained in \#3, in this paper we propose a novel approach enabling researcher in the field to systematically synthesize biological mathematical models to CMOS circuitry operating in strong--inversion. To the best of our knowledge, this is the first systematic strong--inversion circuit capable of emulating such nonlinear bilateral dynamical systems. The application of the method is verified by synthesizing a relatively complex neuron model and transistor--level simulations confirm that the resulting circuits are in good agreement with their mathematical counterparts. Further application of the proposed circuitry on different case studies is left to the interested readers.

\begin{figure}[ht]
\vspace{-20pt}
\normalsize
\centering
\includegraphics[trim = 0in 0in 0in 0in, clip, height=2.8in]{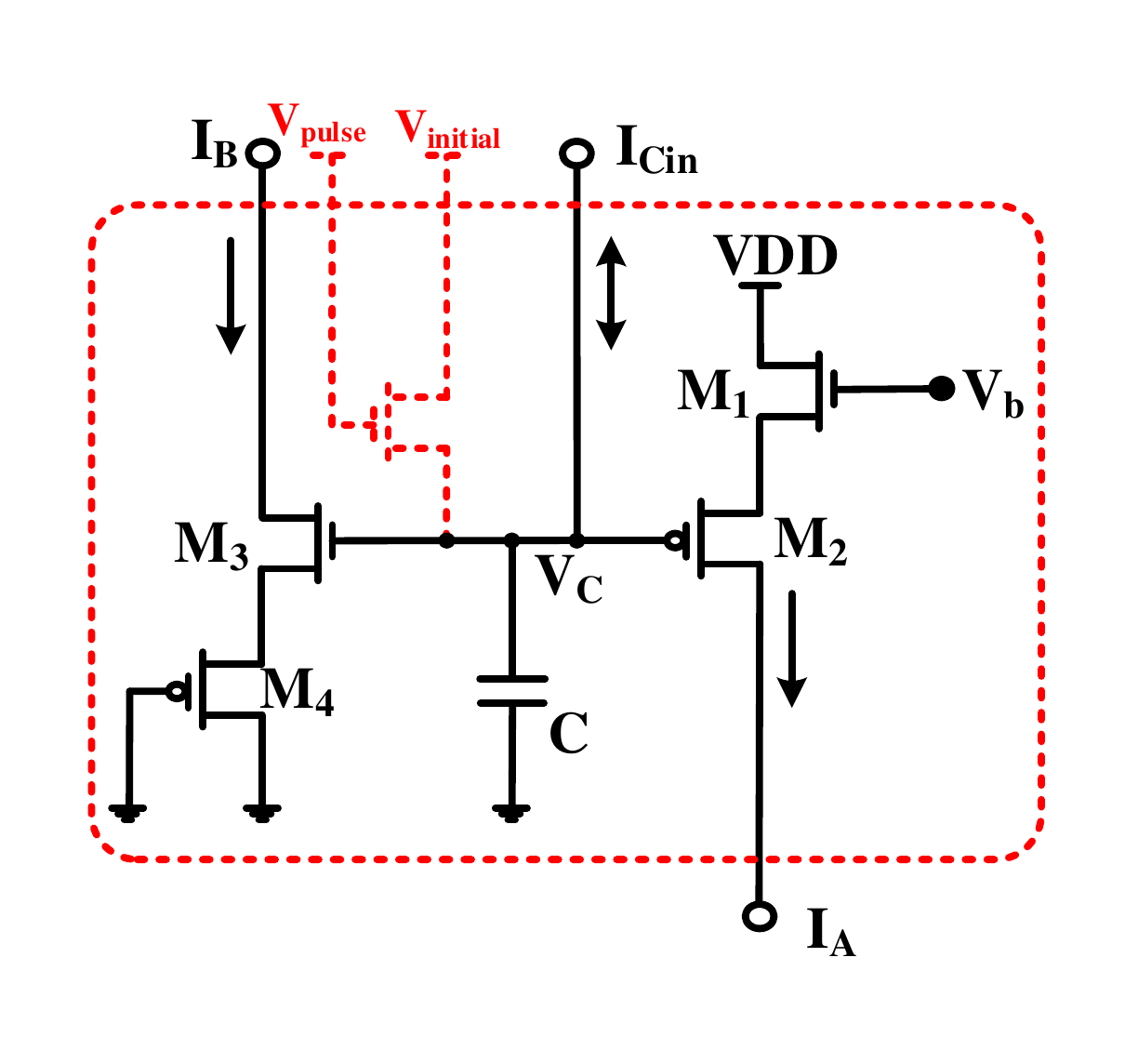}
\vspace{-5pt}
\caption{The ``main core" including the initialization circuit highlighted with red color.}
\label{fig:51_1}
\end{figure}

\section{A Novel Strong--inversion CMOS Circuitry}
In this section, a novel current--input current--output circuit is proposed that supports a systematic realization procedure of strong--inversion circuits capable of computing bilateral dynamical systems at higher speed compared to the previously proposed log--domain circuit. The validity of our approach is verified by nominal simulated results with realistic process parameters from the commercially available AMS 0.35 $\mu m$ technology. The current relationship of an NMOS and PMOS transistor operating in strong--inversion saturation when $\lvert V_{DS}\rvert>\lvert V_{GS}\rvert-\lvert V_{th}\rvert$ can be expressed as follows:
\begin{equation}\label{eq:51_1}
I_{D_n}=\frac{1}{2}\mu_n C_{ox}(\frac{W}{L})_n(V_{GS}-V_{th})^2
\end{equation}

\begin{equation}\label{eq:51_2}
I_{D_p}=\frac{1}{2}\mu_p C_{ox}(\frac{W}{L})_p(V_{SG}-V_{th})^2
\end{equation}

where $\mu_n$ and $\mu_p$ are the charge--carrier effective mobility for NMOS and PMOS transistors, respectively; $W$ is the gate width, $L$ is the gate length, $C_{ox}$ is the gate oxide capacitance per unit area and $V_{th}$ is the threshold voltage of the device.

\par Setting $k_{n}=\frac{1}{2}\mu_n C_{ox}(\frac{W}{L})_n$ and $k_{p}=\frac{1}{2}\mu_p C_{ox}(\frac{W}{L})_p$ in (\ref{eq:51_1}) and (\ref{eq:51_2}) and differentiating with respect to time, the current expression for $I_A$ (see Figure \ref{fig:51_1}) yields:

\begin{equation}\label{eq:51_3}
\dot{I}_A=\overbrace{2k_n(V_{GS}-V_{th})}^{\sqrt {k_nI_A}}\dot{V}_{GS_1}
\end{equation}

\begin{equation}\label{eq:51_4}
\dot{I}_A=\overbrace{2k_p(V_{SG}-V_{th})}^{\sqrt{k_pI_A}}\dot{V}_{SG_2}
\end{equation}

(\ref{eq:51_3}) and (\ref{eq:51_4}) are equal, therefore:
\begin{equation}\label{eq:51_5}
\dot{V}_{SG_2}=\sqrt{\frac{k_n}{k_p}}\dot{V}_{GS_1}=\beta \dot{V}_{GS_1}
\end{equation}
where $\beta=\sqrt{\frac{k_n}{k_p}}$. Similarly, we can derive the following equation for transistors $M_3$ and $M_4$:
\begin{equation}\label{eq:51_6}
\dot{V}_{SG_4}=\sqrt{\frac{k_n}{k_p}}\dot{V}_{GS_3}=\beta \dot{V}_{GS_3}.
\end{equation}
\par The application of Kirchhoff's Voltage Law (KVL) and applying the derivative function show the following relations:
 \begin{equation}\label{eq:51_7}
\dot{V}_{C}=-(\dot{V}_{GS_1}+\dot{V}_{SG_2})
\end{equation}

  \begin{equation}\label{eq:51_8}
\dot{V}_{C}=+(\dot{V}_{GS_3}+\dot{V}_{SG_4})
\end{equation}
 where $V_C$ is the capacitor voltage and $V_b$ the bias voltage which is constant (see Figure \ref{fig:51_1}). Substituting (\ref{eq:51_5}) and (\ref{eq:51_6}) into (\ref{eq:51_7}) and (\ref{eq:51_8}) respectively yields:

 \begin{equation}\label{eq:51_9}
\dot{V}_{C}=-\dot{V}_{GS_1}\cdot(1+\beta)
\end{equation}

\begin{equation}\label{eq:51_10}
\dot{V}_{C}=+\dot{V}_{GS_3}\cdot(1+\beta).
\end{equation}

Setting the current $I_{out}=I_B-I_A$ in Figure \ref{fig:51_1} as the state variable of our system and using (\ref{eq:51_3}) and the corresponding equation for $I_B$, the following relation is derived:

\begin{figure}[t]
\begin{subfigure}{.5\textwidth}
  \centering
\includegraphics[trim = 0in 0in 0in 0in, clip, height=2.3in]{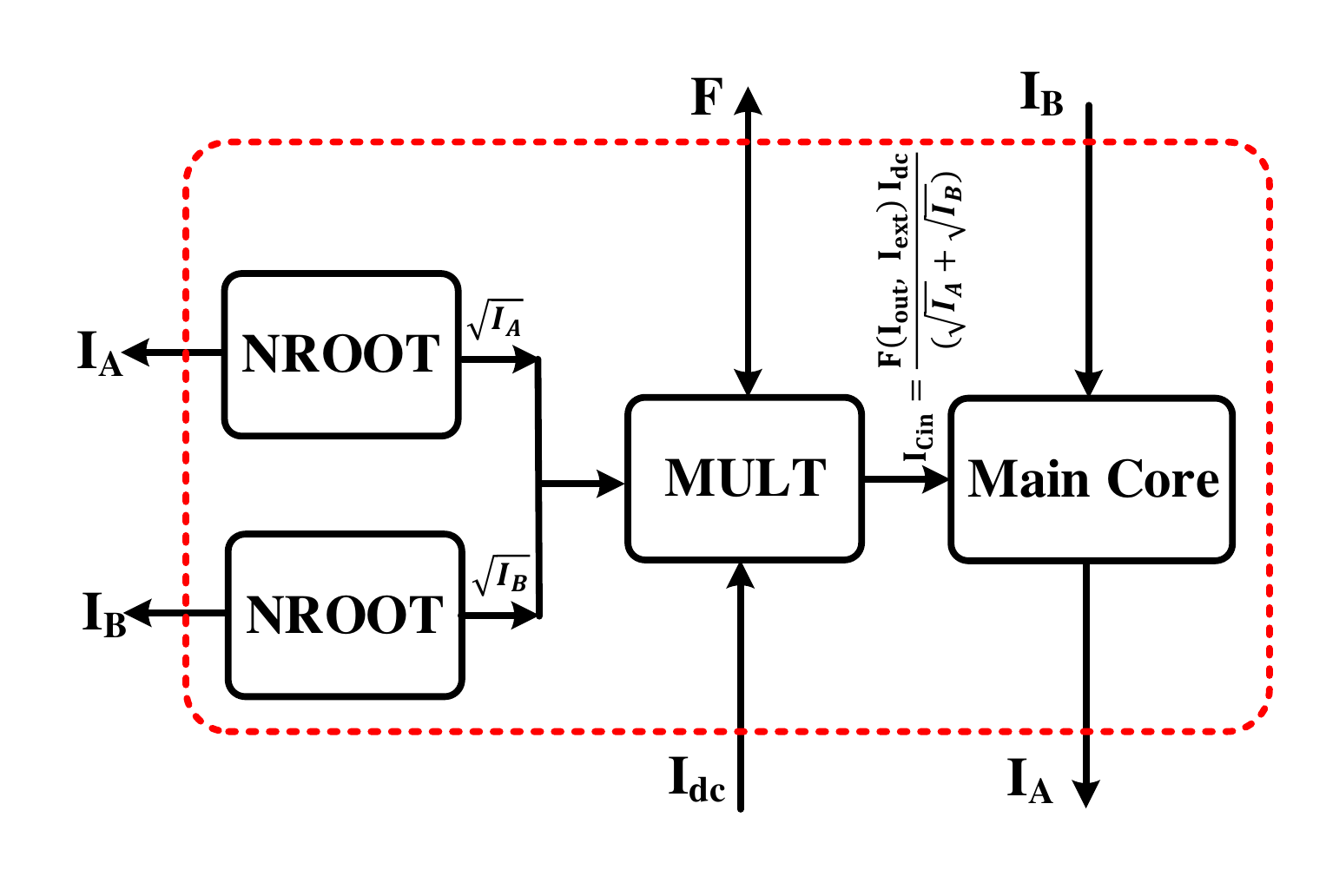}
\label{fig:a}
\end{subfigure}%
\begin{subfigure}{.5\textwidth}
  \centering
\includegraphics[trim = 0in 0in 0in 0in, clip, height=2.5in]{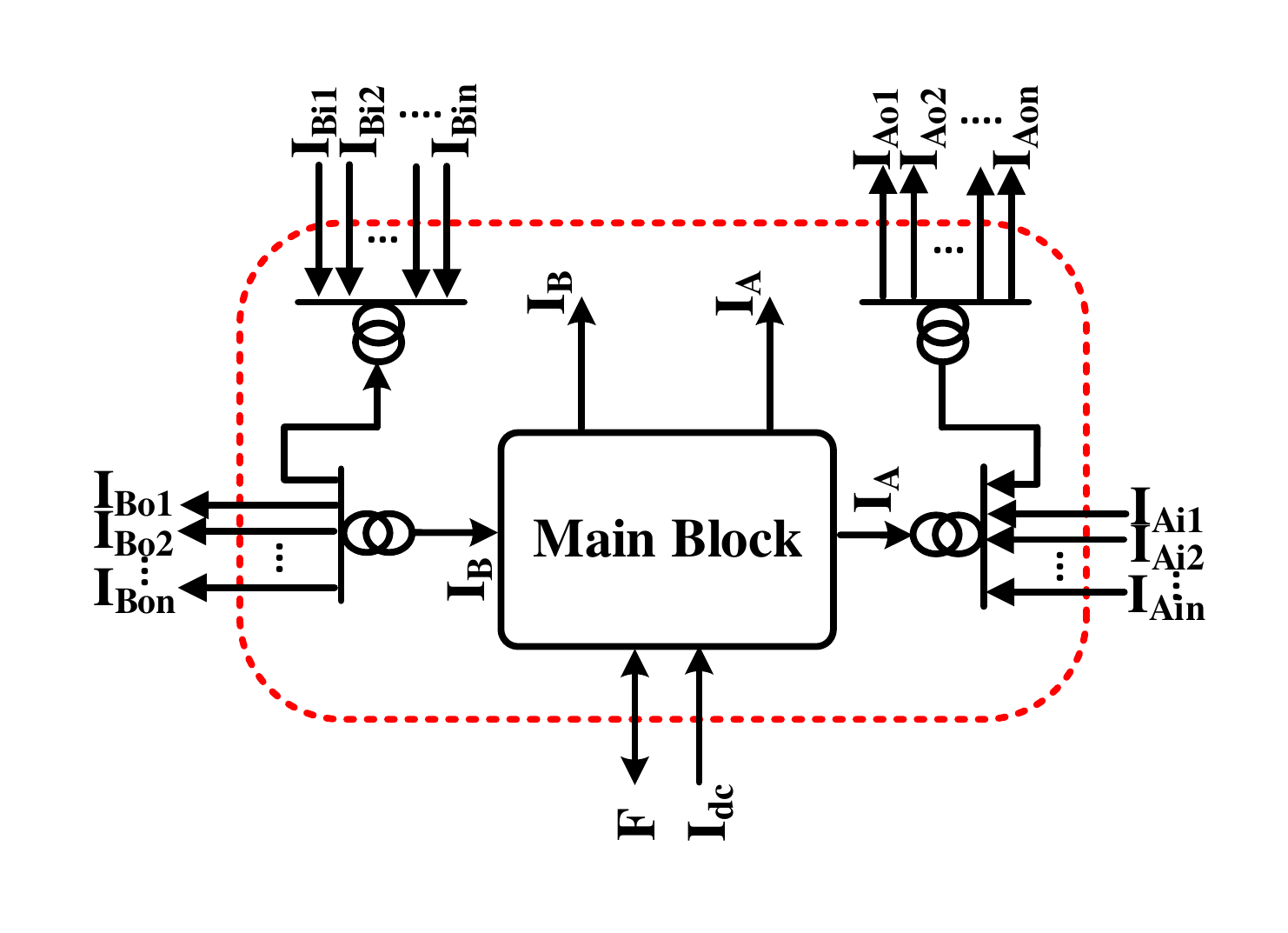}
\label{fig:b}
\end{subfigure}
  \caption{(a) The ``main block" including the main core and two current--mode root square blocks and a bilateral multiplier. (b) The final high speed circuit including the ``main block" with several copied currents (the current mirrors are represented with double circle symbols)}
\label{fig:51_2}
\end{figure}

\begin{equation}\label{eq:51_11}
\dot{I}_{out}=\dot{I}_B-\dot{I}_A=2\sqrt{k_nI_B}\dot{V}_{GS_3}-2\sqrt{k_pI_A}\dot{V}_{GS_1}
\end{equation}

by substituting (\ref{eq:51_9}) and (\ref{eq:51_10}) in (\ref{eq:51_11}):
\begin{equation}\label{eq:51_12}
\dot{I}_{out}=(\sqrt{I_A}+\sqrt{I_B})\cdot \frac{2\sqrt{k_n}\dot{V}_C}{2+\beta}.
\end{equation}
\par Bearing in mind that the capacitor current $I_{Cin}$ can be expressed as $C\dot{V}_C$, relation (\ref{eq:51_12}) yields:
\begin{equation}\label{eq:51_13}
\dot{I}_{out}=(\sqrt{I_A}+\sqrt{I_B})\cdot \frac{2\sqrt{k_n}I_{Cin}}{(2+\beta)C}.
\end{equation}

\par One can show that:
\begin{equation}\label{eq:51_14}
\frac{(2+\beta)C}{2\sqrt{k_n}\cdot I_{dc}}\dot{I}_{out}=\frac{(\sqrt{I_A}+\sqrt{I_B})}{I_{dc}}\cdot I_{Cin}.
\end{equation}
\par Equation (\ref{eq:51_14}) is the main core's relation. In order for a high speed mathematical dynamical system with the following general form to be mapped to (\ref{eq:51_14}):
\begin{equation}\label{eq:51_15}
\tau\dot{I}_{out} =F(I_{out}, I_{ext})
\end{equation}
where $I_{ext}$ and $I_{out}$ are the external and state variable currents, the quantities $\frac{C}{I_{dc}}$ and $I_{Cin}$ must be respectively equal to $\frac{2\tau\sqrt{k_n}}{(2+\beta)}$ and $\frac{F(I_{out}, I_{ext})I_{dc}}{(\sqrt{I_A}+\sqrt{I_B})}$. Note that the ratio value $\frac{C}{I_{dc}}$ can be satisfied with different individual values for $C$ and $I_{dc}$. These values should be chosen appropriately according to practical considerations (see Section V.G). Since $F$ is a bilateral function, in general, it will hold:
\begin{equation}\label{eq:51_16}
I_{Cin}=\overbrace{\frac{F^+(I_A,I_B,I_{ext}^+,I_{ext}^-)I_{dc}}{(\sqrt{I_A}+\sqrt{I_B})}}^{I_{Cin}^+}-\overbrace{\frac{F^-(I_A,I_B,I_{ext}^+,I_{ext}^-)I_{dc}}{(\sqrt{I_A}+\sqrt{I_B})}}^{I_{Cin}^-}
\end{equation}
where $I_{Cin}^+$ and $I_{Cin}^-$ are calculated respectively by a root square block (see Figure \ref{fig:51_2}(a) and $I_{ext}$ is separated to + and -- signals by means of splitter blocks. Note that $I_{dc}$ is a scaling dc current and $\tau$ has dimensions of $second(s)$. Since $I_{Cin}$ can be a complicated nonlinear function in dynamical systems, we need to provide copies of $I_{out}$ or equivalently of $I_A$ and $I_B$ to simplify the systematic computation at the circuit level. Therefore, the higher hierarchical block shown in Figure \ref{fig:51_2}(b) is defined as the NBDS (Nonlinear Bilateral Dynamical System) circuit \cite{16} (see Figure Figure \ref{fig:51_2}(b)) including the main block and associated current mirrors. The form of (\ref{eq:51_15}) is extracted for a 1--D dynamical system and can be extended to $N$ dimensions in a straightforward manner as follows:
\begin{equation}\label{eq:51_17}
\tau_N\dot{I}_{out_N} =F_N(\bar{I}_{out},\bar{I}_{ext})
\end{equation}
where $\frac{C_N}{I_{dc_N}}=\frac{2\tau_N\sqrt{k_n}}{(2+\beta)}$ and $I_{Cin_N}=\frac{F_N(\bar{I}_{out},~\bar{I}_{ext})I_{dc_N}}{(\sqrt{I_{A_N}}+\sqrt{I_{B_N}})}$.

\begin{figure}[t]
\begin{subfigure}{.5\textwidth}
  \centering
\includegraphics[trim = 0in 0in 0in 0in, clip, height=3in]{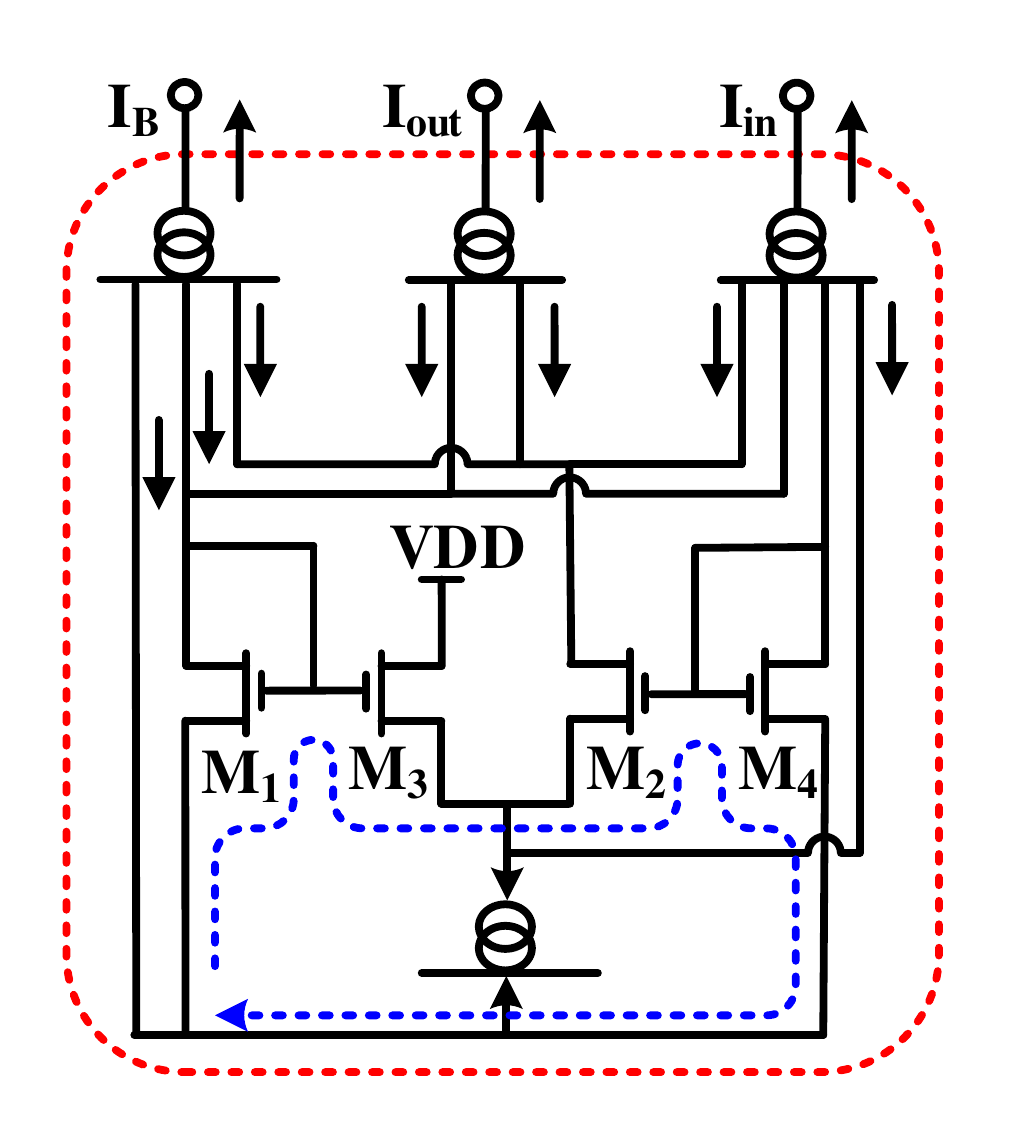}
  \label{fig:sfig1}
\end{subfigure}%
\begin{subfigure}{.5\textwidth}
  \centering
\includegraphics[trim = 0in 0in 0in 0in, clip, height=3in]{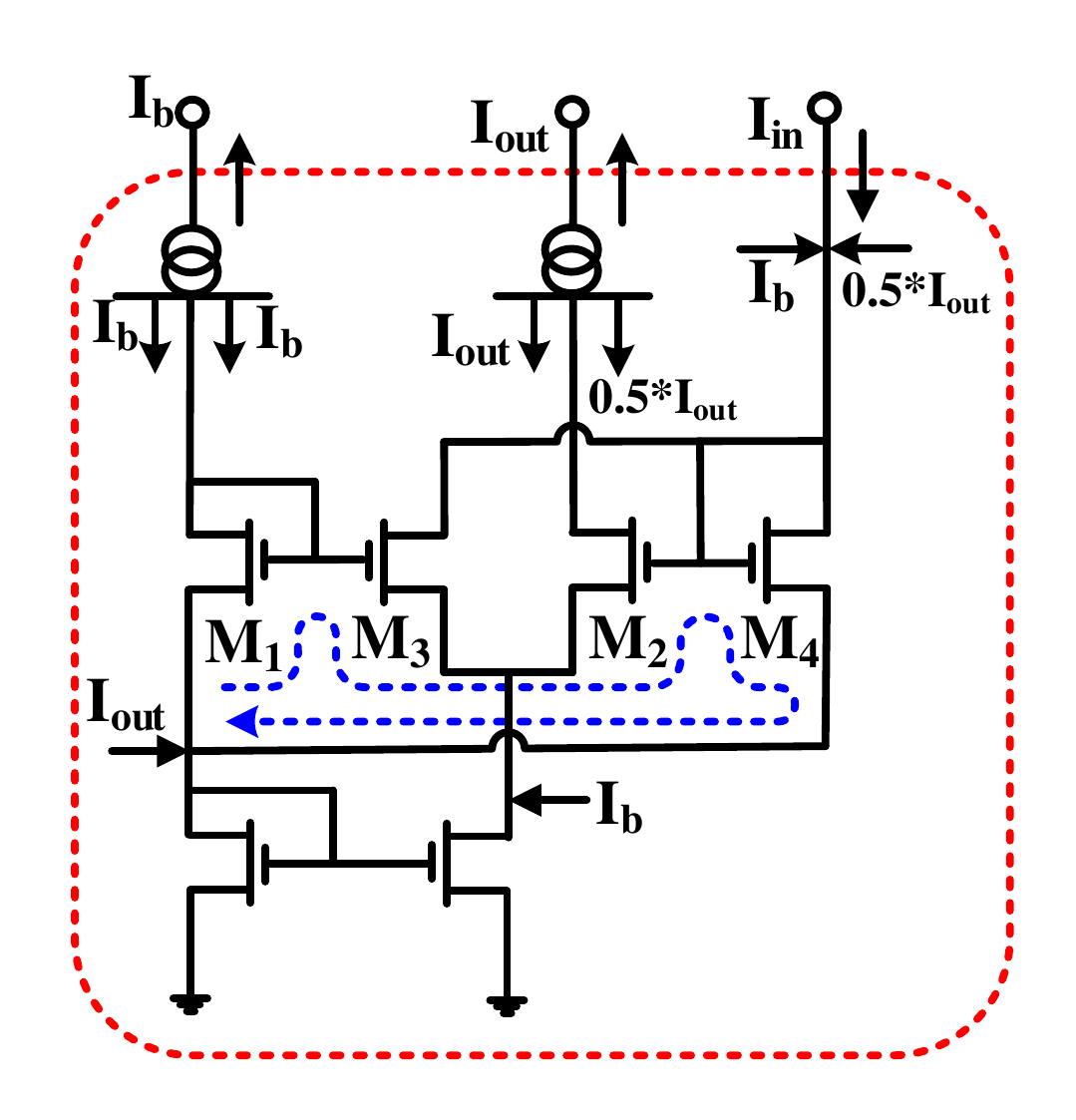}
  \label{fig:sfig2}
\end{subfigure}
  \caption{(a) Transistor level representation of the basic Root Square block. The current mirrors are represented with double circle symbols. (b) Transistor level representation of the MULT core block. The current mirrors are represented with double circle symbols.)}
\label{fig:51_3}
\end{figure}

\section{Basic Electrical Blocks}
\subsection{Root Square Block}
This block performs current mode root square function on single--sided input signals. By setting $(\frac{W}{L})_{1\&2}=4\times(\frac{W}{L})_{3\&4}$, considering $I_1, I_2, I_3$ and $I_4$ as the currents flowing respectively into $M_1, M_2, M_3$ and $M_4$ and all transistors operate in strong--inversion saturation, the governing TL principle for this block becomes (highlighted with dotted blue arrow):
\begin{equation}\label{eq:51_18}
\frac{1}{2}(\sqrt{I_1}+\sqrt{I_2})=\sqrt{I_3}+\sqrt{I_4}
\end{equation}
\par By pushing specific currents (copied by current mirrors) according to Figure \ref{fig:51_3} (a) into the TL's transistors we have:
\begin{equation}\label{eq:51_19}
\begin{cases}
I_1=I_2=I_{in}+I_{out}+I_b\\
I_3=I_b,~I_4=I_{in}
\end{cases}
\end{equation}
Substituting (\ref{eq:51_19}) into (\ref{eq:51_18}) yields:
\begin{equation}\label{eq:51_20}
\frac{1}{2}\times(\sqrt{I_{in}+I_{out}+I_b}+\sqrt{I_{in}+I_{out}+I_b}=\sqrt{I_{in}}+\sqrt{I_b}
\end{equation}
By squaring both sides of (\ref{eq:51_20}):
\begin{equation}\label{eq:51_21}
I_{in}+I_{out}+I_b=\sqrt{I_{in}}+\sqrt{I_{in}\cdot I_b}
\end{equation}
and finally:
\begin{equation}\label{eq:51_22}
I_{out}=2\sqrt{I_{in}\cdot I_b}
\end{equation}

\subsection{MULT Core Block}
This block is the main core forming the final bilateral multiplier which is introduced in the next subsection. The block contains six transistors as well as two current mirrors. By assuming $I_1, I_2, I_3$ and $I_4$ as the currents flowing respectively into $M_1, M_2, M_3$ and $M_4$ and the same $\frac{W}{L}$ aspect ratio for all transistors operating in strong--inversion saturation, the KVL at the highlighted TL with dotted blue arrow yields:
\begin{equation}\label{eq:51_23}
\sqrt{I_1}+\sqrt{I_2}=\sqrt{I_3}+\sqrt{I_4}
\end{equation}
\par By forcing specific currents (copied by current mirrors) according to Figure \ref{fig:51_3} (b) into the TL's transistors we have:
\begin{equation}\label{eq:51_24}
\begin{cases}
I_1=I_b,~I_2=I_{out}\\
I_3=I_4=\frac{1}{2}(I_{in}+\frac{I_{out}}{2}+I_b)
\end{cases}
\end{equation}
Substituting (\ref{eq:51_24}) into (\ref{eq:51_23}) yields:
\begin{equation}\label{eq:51_25}
\sqrt{I_{out}}+\sqrt{I_b}=2\sqrt{\frac{1}{2}(I_{in}+\frac{I_{out}}{2}+I_b)}
\end{equation}
By squaring both sides of (\ref{eq:51_25}):
\begin{equation}\label{eq:51_26}
\sqrt{I_{out}\cdot I_b}=I_{in}+\frac{1}{2}I_b
\end{equation}
and:
\begin{equation}\label{eq:51_27}
I_{out}=\frac{(I_{in}+\frac{1}{2}I_b)^2}{I_b}
\end{equation}

\subsection{Bilateral MULT Block}
This block is able to perform current mode multiplication operation on bilateral input signals. If inputs are split to positive and negative sides we have:
\begin{equation}\label{eq:51_28}
\begin{cases}
X=X^+-X^-\\
Y=Y^+-Y^-.
\end{cases}
\end{equation}
\par The multiplication result can be expressed as $XY=X^+Y^++X^-Y^--(X^-Y^++Y^-X^+)$. By extending equation (\ref{eq:51_27}) to $\frac{I_{in}^2}{I_b}+\frac{I_b}{4}+I_{in}$ for every basic MULT core block, the output signal constructed by a positive and negative side can be written as:
\begin{multline}
I_{out}=\frac{(X^++Y^+)^2}{I_b}+(X^++Y^+)+\frac{I_b}{4}+\frac{(X^-+Y^-)^2}{I_b}+(X^-+Y^-)+\frac{I_b}{4}\\
-\frac{(X^-+Y^+)^2}{I_b}-(X^-+Y^+)-\frac{I_b}{4}-\frac{(X^++Y^-)^2}{I_b}-(X^++Y^-)-\frac{I_b}{4}
\end{multline}
and by further simplifications:
\begin{equation}\label{eq:51_29}
I_{out}=\overbrace{\frac{2(X^+Y^++X^-Y^-)}{I_b}}^{I_{out}^+}-\overbrace{\frac{2(X^-Y^++X^+Y^-)}{I_b}}^{I_{out}^-}=\frac{2XY}{I_b}
\end{equation}

\begin{figure*}[t]
\vspace{-20pt}
\normalsize
\centering
\includegraphics[trim = 0in 0in 0in 0in, clip, height=3.2in]{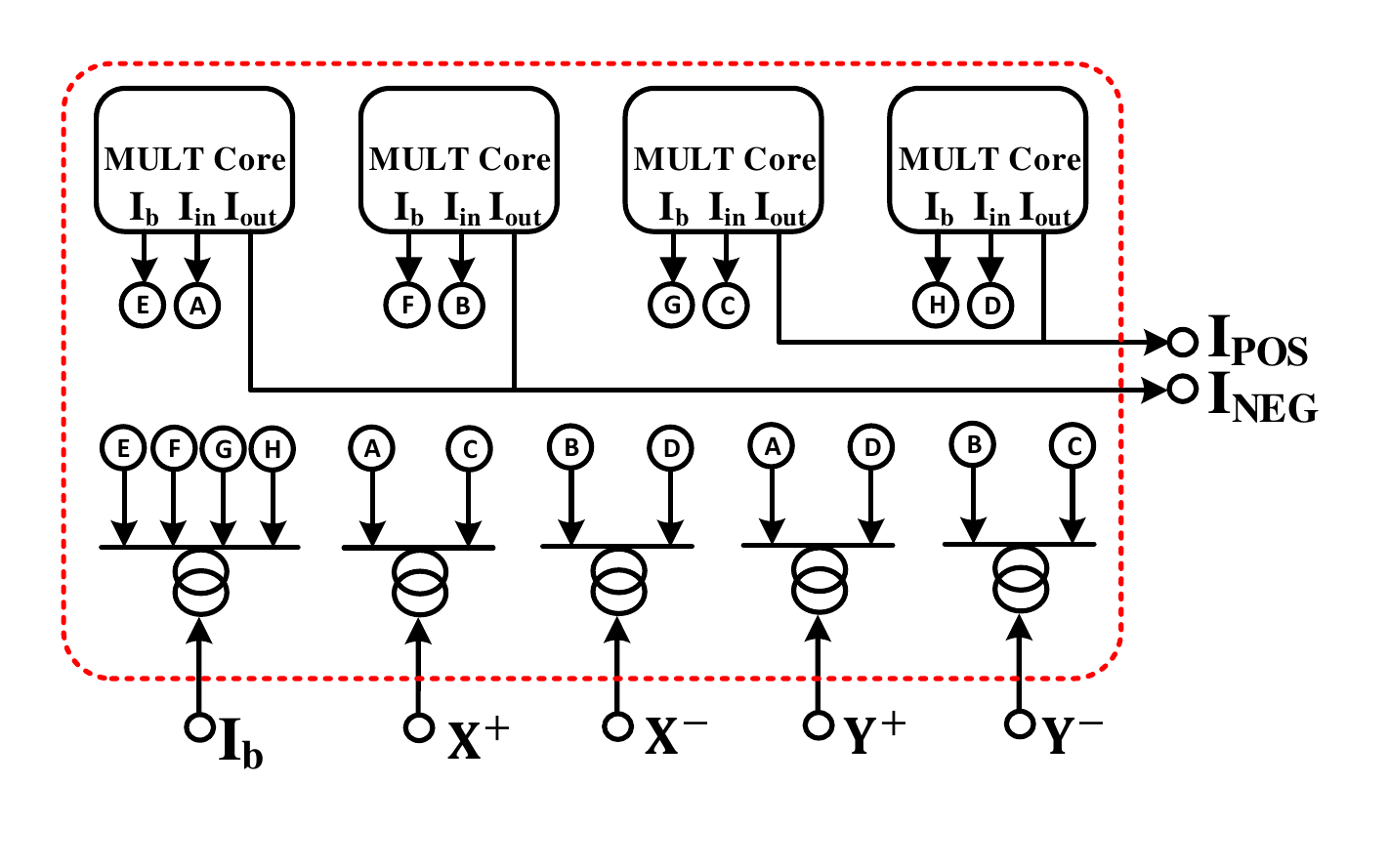}
\vspace{-5pt}
\caption{Schematic and symbolic representation of the bilateral MULT block comprising current mirrors and MULT Core block.}
\label{fig:51_6}
\end{figure*}

\subsection{Circuit Realization of FHN neuron model}
The systematic synthesis procedure provides the flexibility and convenience required for the realization of nonlinear dynamical systems by computing their time-dependent dynamical behavior. In this subsection, we showcase the methodology through which we systematically map the mathematical dynamical models onto the proposed electrical circuit. Here, the application of the method is demonstrated by synthesizing the 2--D nonlinear FitzHugh–Nagumo neuron model. In the FHN neuron model \cite{25} with the following representation: $\dot{v}=v-\frac{v^3}{3}-w+I_{ext}$ and $\dot{w}=0.18(v+0.7-0.8w)$ describing the membrane potential's and the recovery variable's velocity, the state variables in the absence of input stimulation remain at $(v, w)\approx(-1.2, -0.6)$, while these values go up to $(v, w)\approx(2, 1.7)$ in the presence of input stimulation.  According to this biological dynamical system, we can start forming the electrical equivalent using (\ref{eq:51_17}):
\begin{equation}\label{eq:51_30}
\begin{cases}
\frac{(2+\beta)C}{2\sqrt{k_n}\cdot I_{dc_v}}\dot{I}_{out_v}=F_v(I_{out_v},I_{out_w},I_{ext})\\
\frac{(2+\beta)C}{2\sqrt{k_n}\cdot I_{dc_w}}\dot{I}_{out_w}=F_w(I_{out_v},I_{out_w})
\end{cases}
\end{equation}
where $I_{dc_v}=80nA$, $I_{dc_w}=a\cdot I_{dc_v}=6.4nA$, $F_v$ and $F_w$ are functions given by:
\begin{equation}\label{eq:5_33}
\begin{cases}
F_v(I_{out_v},I_{out_w},I_{ext})=I_{out_v}-\frac{I_{out_v}^3}{I_bI_x}-I_{out_w}+I_{ext}\\
F_w(I_{out_v},I_{out_w})=(I_{out_v}+I_{c}-\frac{I_dI_{out_w}}{I_x})
\end{cases}
\end{equation}
where $I_b=3 uA$, $I_c=0.7 uA$, $I_d=0.8 uA$ and $I_x=1 uA$.

\begin{figure*}[t]
\vspace{-20pt}
\normalsize
\centering
\includegraphics[trim = 0.1in 0.1in 0.1in 0.1in, clip, width=6.5in]{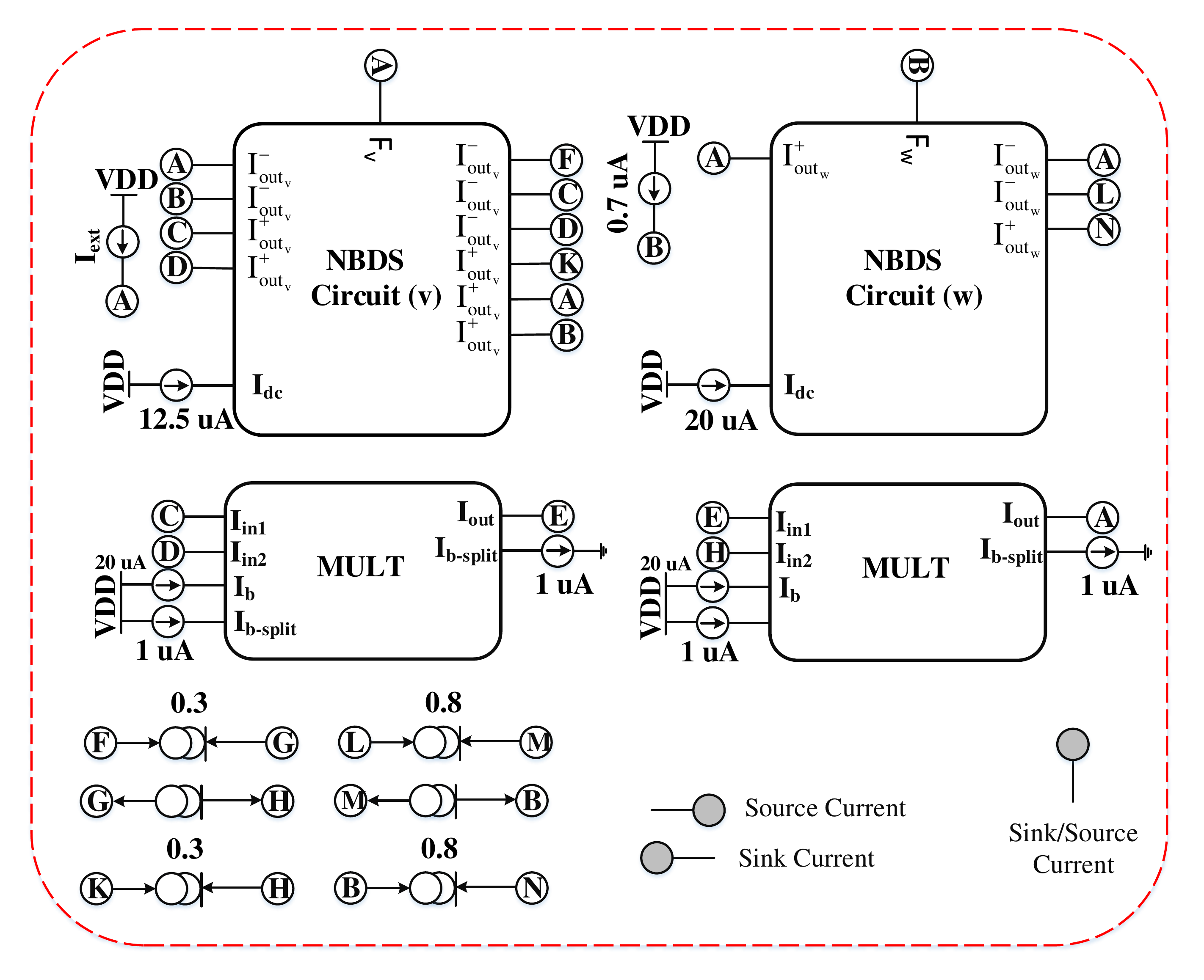}
\vspace{-5pt}
\caption{A block representation of the total circuit implementing the 2--D FHN neuron model.}
\vspace{-15pt}
\label{fig:51_7}
\end{figure*}

\par Schematic diagrams for the FHN neuron model is seen in Figure \ref{fig:51_7}, including the symbolic representation of the basic TL blocks
introduced in the previous sections. According to these diagrams, it is observed how the mathematical model is mapped onto the proposed electrical circuit. The schematic contains two NBDS circuits implementing the two dynamical variables, followed by two MULT and current mirrors realizing the dynamical functions. As shown in the figure, according to the neuron model, proper bias currents are selected and the correspondence between the biological voltage and electrical current is $V\iff uA$.

\begin{figure*}[t]
\vspace{-20pt}
\normalsize
\centering
\includegraphics[trim = 0.1in 0.1in 0.1in 0.1in, clip, width=6in]{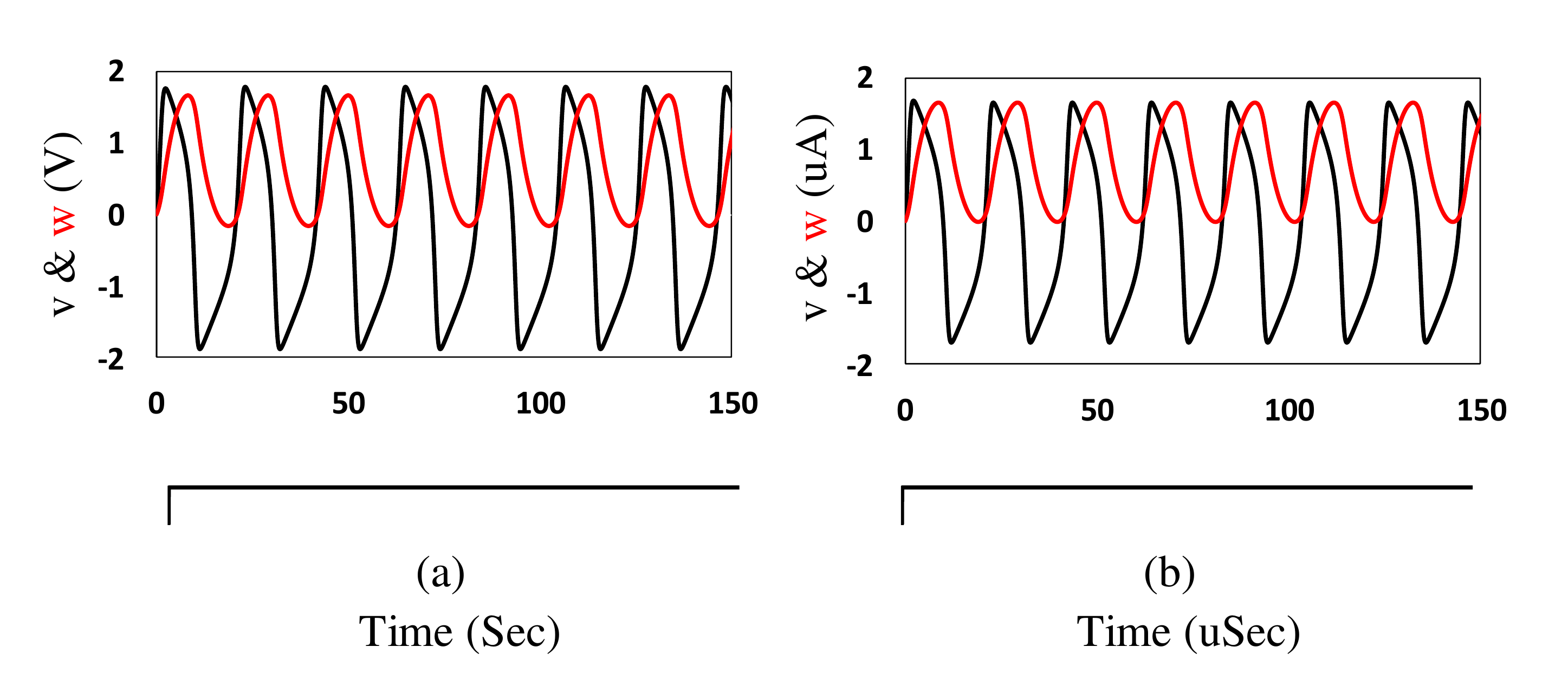}
\vspace{-5pt}
\caption{Time-domain representations of regular spiking for (a) for MATLAB and (b) Cadence respectively.}
\label{fig:51_8}
\end{figure*}

\begin{table}[t]
\caption{Electrical Parameter Values for the Simulated FHN Neuron Model operating in strong--inversion.}   
\centering          
\begin{tabular}{c c}    
\hline\hline                        
Specifications & Value\\ [0.5ex]  
\hline                      
Power Supply (Volts)& 3.3\\
Bias Voltage (Volts)& 3.3\\
Capacitances (pF)& 800\\
$\frac{W}{L}$ ratio of PMOS and NMOS Devices ($\frac{\mu m}{\mu m}$)& $\frac{12}{1}$ and $\frac{10}{1}$\\
Static Power Consumption ($m W$)& 8.94\\
\hline          
\end{tabular}
\label{table:51_1}    
\end{table}

\section{Discussion}
\par Here, we demonstrate the simulation--based results of the high speed circuit realization of the FHN neuron model. The hardware results simulated by the Cadence Design Framework (CDF) using the process parameters of the commercially available AMS 0.35 $\mu m$ CMOS technology are validated by means of MATLAB simulations as shown in Figure \ref{fig:51_8}. For the sake of frequency comparison, a regular spiking mode is chosen. Generally, results confirm an acceptable compliance between the MATLAB and Cadence simulations while the hardware model operates at higher speed (almost 1 million times faster than real--time). Table \ref{table:51_1} summarizes the specifications of the proposed circuit applied to this case study. As shown in the table, the circuit uses a higher $V_b$ compared to the subthreshold version to force the circuit to operate in strong--inversion region. This comes at the expense of higher power consumption (95000 times higher than the subthreshold version). 

\bibliographystyle{unsrt}	
\bibliography{paper}

\end{document}